\documentclass[3p,times,preprint,11.5pt]{elsarticle}

\usepackage{amssymb}
\usepackage{amsmath}
\usepackage{multirow, multicol}
\usepackage{hyperref}
\usepackage{tabularray}

\hypersetup{
    colorlinks=true, 
    urlcolor=magenta,   
}

\journal{Advanced Engineering Informatics}

\begin{document}

\begin{frontmatter}



\title{Integrating Frequency Guidance into
Multi-source Domain Generalization for Bearing
Fault Diagnosis}


\author[label1,label2]{Xiaotong Tu\corref{cor1}}
\ead{xttu@xmu.edu.cn}
\author[label1,label2]{Chenyu Ma\corref{cor1}}
\ead{machenyu@stu.xmu.edu.cn}

\author[label1,label3]{Qingyao Wu}
\author[label1,label3]{Yinhao Liu}

\author[label1,label2]{Hongyang Zhang\corref{cor1}\footnote{Corresponding author.}}
\ead{hyzhang@stu.xmu.edu.cn}

\cortext[cor1]{All the authors contribute equally to this work.}


\affiliation[label1]{organization={Key Laboratory of Multimedia Trusted Perception and Efficient Computing, Ministry of Education of China},
            addressline={Xiamen University},
            city={Xiamen},
            postcode={361005},
            state={Fujian},
            country={China}}

\affiliation[label2]{organization={School of Informatics},
            addressline={Xiamen University},
            city={Xiamen},
            postcode={361005},
            state={Fujian},
            country={China}}

\affiliation[label3]{organization={Institute of Artificial Intelligence},
            addressline={Xiamen University},
            city={Xiamen},
            postcode={361005},
            state={Fujian},
            country={China}}

\begin{abstract}
Recent generalizable fault diagnosis researches have effectively tackled the distributional shift between unseen working conditions. Most of them mainly focus on learning domain-invariant representation through feature-level methods. However, the increasing numbers of unseen domains may lead to domain-invariant features contain instance-level spurious correlations, which impact the previous models’ generalizable ability. To address the limitations, we propose the Fourier-based Augmentation Reconstruction Network, namely FARNet. The methods are motivated by the observation that the Fourier phase component and amplitude component preserve different semantic information of the signals, which can be employed in domain augmentation techniques. The network comprises an amplitude spectrum sub-network and a phase spectrum sub-network, sequentially reducing the discrepancy between the source and target domains. To construct a more robust generalized model, we employ a multi-source domain data augmentation strategy in the frequency domain. Specifically, a Frequency-Spatial Interaction Module (FSIM) is introduced to handle global information and local spatial features, promoting representation learning between the two sub-networks. To refine the decision boundary of our model output compared to conventional triplet loss, we propose a manifold triplet loss to contribute to generalization. Through extensive experiments on the CWRU and SJTU datasets, FARNet demonstrates effective performance and achieves superior results compared to current cross-domain approaches on the benchmarks. Code will be available at: https://github.com/HRT00/FAR-Net.
\end{abstract}




\end{frontmatter}



\section{Introduction}
\label{sec:introduction}
Intelligent fault diagnosis technologies, grounded in artificial intelligence methodologies, have experienced significant advancements, facilitating the precise and efficient automated monitoring of industrial equipment health status \cite{r1,r2}. The strategic deployment of this technology is an indispensable condition for realizing industrial intelligence.

\begin{figure*}[ht]
\centerline{\includegraphics[width=\linewidth]{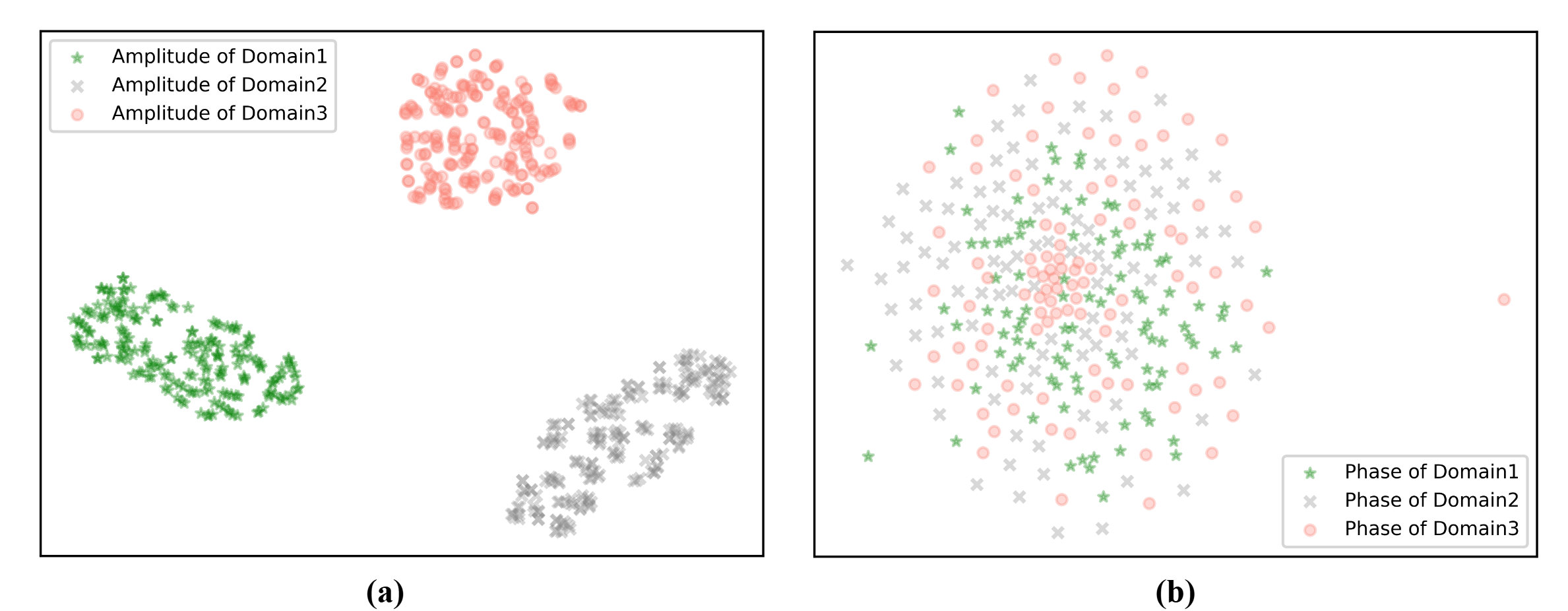}}\caption{The T-SNE\cite{r59} of amplitude and phase of category inner race fault (IRF) with 7 mils, 14 mils, and 21 mils (three domains) on CWRU dataset are depicted on (a) and (b), respectively. It's obvious that amplitude features are compose of domain-relevant information and phase features are compose of domain-invariant information, which can help to obtain more diverse data for domain generalization.}
\label{fig1}
\end{figure*}

Deep learning-based fault diagnosis approaches \cite{r47, r49, r51} are particularly adept at uncovering the intricate relationships within fault data, thereby achieving promising results in the supervised setting. However, most of them assume consistent data distributions between training and testing data. In real-world industrial applications, factors such as variations in work speed, load, and mechanical equipment inevitably amplify the distribution gap between the data in the source and target domains \cite{r48, r50}. Essentially, collecting adequate training data from all possible working conditions is costly and even impossible in some practical environments. Thus, it's more common that source domain models lack access to unlabeled target domain data during training. And it becomes significant to empower deep learning models with the capability to mitigate domain shift and generalize to unseen data domains, crucial for effective industrial fault diagnosis. To tackle the aforementioned challenges, we propose a domain generalization (DG) method for fault diagnosis, which aims to train robust models in one or more relevant source domains without target domain data, thereby enabling generalization to any unseen target domain. DG methods can be categorized into three types: simulating diverse scenarios within a meta-learning framework \cite{r15, r52}, learning domain-invariant information by aligning multiple different yet related source domains \cite{r11, r54}, and transforming source domain data through data augmentation techniques \cite{r16, r53}.

\begin{figure*}[ht]
\centerline{\includegraphics[width=\linewidth]{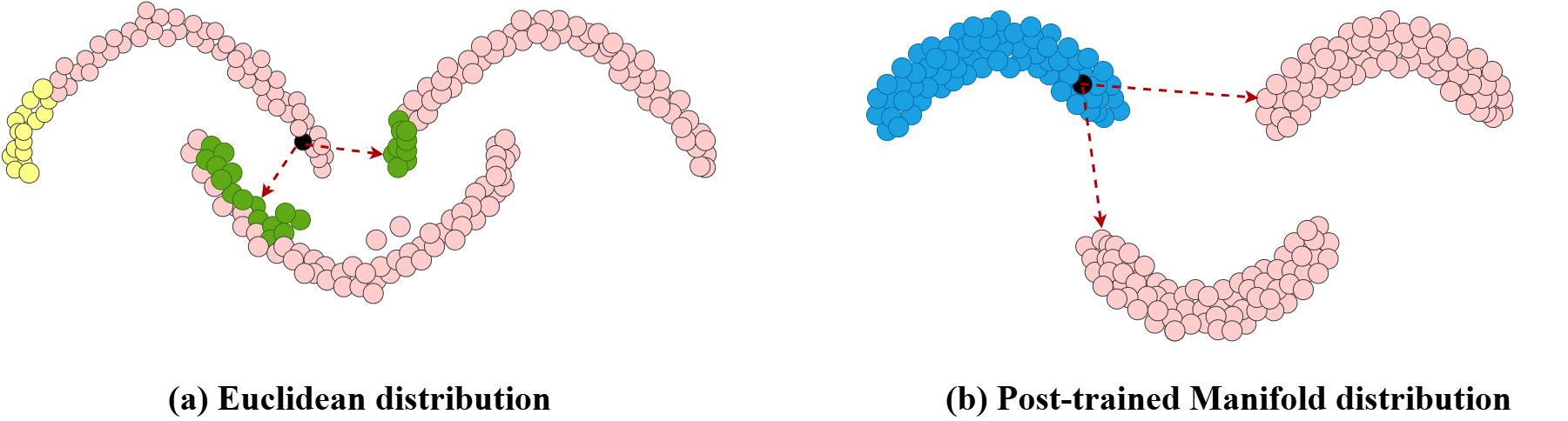}}\caption{The data distribution of a synthetic multi-class dataset in Euclidean space and Manifold space in $\mathcal{R}^{2}$. The original distribution of Euclidean space is illustrated in (a). The black dot denotes the anchor, the green circles represent 
hard-negative samples and the yellow circles are hard-positive samples. A non-linear activation conducted on the Euclidean space to obtain a leaky-relu-like distance measurement, and post-trained Manifold distribution in (b) is obtained through the training of triplet loss with the new measurement. The manifold triplet loss attempts to refine the distribution which results in the intra-class distribution (blue dots) more narrow and the distance between each class larger.}
\label{fig2}
\end{figure*}

Recently, domain augmentation techniques \cite{r30, r31, r43} seek to generate artificial samples for synthetic domains and alleviating over-fitting issues, have achieved wide attention on the fault diagnosis. For example, Huang \textit{et al.} \cite{r3} proposes a Fourier-based deep exposure correction module for image restoration to enhance visual effects. Motivated by the properties of the Fourier transform, the preservation of high-level semantic information in the phase components of the Fourier spectrum, and the insight that amplitude spectrum components contain low-level statistical information \cite{r18}, we employ data reconstruction and enhancement from the perspective of the frequency domain. To verify our core idea, we visualize the amplitudes and phases parts from the same category but different domains within the Case Western Reserve University (CWRU) rolling bearing dataset \cite{r60} through T-SNE \cite{r59}. As shown in Fig.\ref{fig1}, the phase representations exhibit consistent distributions across different domains, whereas the amplitudes show significant variations across domains. This illustrates that the phase component contains more structural information and is less susceptible to the influence of domain shift.

Based on the above analysis, this paper introduces a Fourier-based Augmentation Reconstruction Network (FARNet). In contrast to conventional convolution operations that focus on learning local representations in the spatial domain, our method proposes interaction blocks to incorporate two representations from spatial domain and frequency domain, enabling the learning of more representative features. As depicted in Fig.\ref{fig3}, the network comprises an amplitude spectrum sub-network and a phase spectrum sub-network. The amplitude spectrum sub-network learns amplitude representations to enhance signal diversity and mitigate the effects of domain-shift issues. Simultaneously, the phase spectrum sub-network learns reconstructed phases to represent refined structures. We use data from one of the source domains as an augmentation target to guide the learning of both the amplitude and phase spectrum sub-networks. The global information captured in the frequency domain and the local information derived from the spatial domain are mutually complementary, both the two domains contribute to augmenting the frequency-spatial representations of the models in low-level tasks \cite{r61, r62}. Thus, we also introduce a Frequency-Spatial Interaction Module (FSIM) as illustrated in Fig.\ref{fig4} to further improve the learning of amplitude and phase representations. The FSIM is tailored in two components (amplitude and phase), and each type of FSIM consists of a frequency branch and a spatial branch to handle global information and local spatial features. This approach allows us to fully leverage the frequency-spatial interaction learning strategy, thereby improving the performance of multi-source domain generalization in the context of fault diagnosis applications.

Furthermore, deep metric learning methods have been introduced in the field of bearing fault diagnosis~\cite{r34, r38, r46} to learn discriminative features that suppress intra-class variations and maximize the gap between samples from different classes, thereby demonstrating excellent performance in cross-domain scenarios. In \cite{r37}, the authors propose adversarial domain adaptation for fault diagnosis using triplet loss. Meanwhile, conditional contrastive loss~\cite{r36}, which seeks domain-invariant features from multi-source domains, was introduced in the domain generalization task. However, major existing methods consider neighbors in Euclidean space to obtain relevant training data. Inspired by research~\cite{r40}, positive samples are distant points on the same manifold, while negative examples are nearby points on different manifolds. Thus, we consider mining hard examples in a manifold space through a non-linear activation. Accordingly, we attempt to propose an improved triplet loss in manifold sphere rather than in Euclidean space which allows the model to find more hard-negative and hard-positive samples and achieve better performance on recognizing the fault categories. As shown in Fig.\ref{fig2} (b), our method can refine the decision boundary and narrow the data distribution for each category in the post-trained manifold space. This approach allows our method to effectively capture the geometrical and structural information in bearing fault signals, encouraging the model to learn more discriminative capabilities in unseen working conditions. To effectively mine the hard negative samples in the manifold space, we design a leaky-relu-like distance measurement, introducing nonlinear activation to exploit more precise category boundaries in multiple source domains.

Our main contributions to this study are as follows:

1. We introduce a novel approach which reconstructs various fault domains through a Frequency-based domain augmentation. We design the Fourier-based Augmentation Reconstruction Network (FARNet), which progressively reconstructing and improving on both amplitude and phase representations.

2. We propose a Frequency-Spatial Interaction Module (FSIM) in two components (amplitude and phase) to effectively integrate global information and local spatial features.

3. We design an innovative manifold triplet loss, aiming to minimize intra-class feature distances and maximize inter-class distances. This leads to more precise decision boundaries, enhancing the overall discriminative capabilities of the model.

4. Extensive experiments under unseen working conditions are conducted using two motor bearing fault datasets, and the results demonstrate that our proposed FARNet outperforms the other state-of-the-art approaches on different DG tasks.

\section{Related Work}
\subsection{Domain Generalization for Fault Diagnosis}
Domain Adaptation (DA) methods have achieved results comparable to supervised learning, but these approaches require data from the target domain, which still presents a significant discrepancy from real-world scenarios. Domain Generalization aims to distill knowledge from source domains and apply it to unseen target domains. To extend diagnostic knowledge to unseen domains, DG methods\cite{r45, r55, r56} have been preliminarily deployed in the context of bearing fault diagnosis. For instance, Zhang \textit{et al.} \cite{r9} proposed a DG method utilizing conditional adversarial training to tackle distribution changes in the target domain, yielding high fault diagnosis rates. Li \textit{et al.} \cite{r4} achieved enhanced generalization through adversarial domain-enhanced training, facilitating the learning of general and enhanced features, which leads to improve model's generalization performance. For domain augmentation method, Zhao \textit{et al.} \cite{r46} used a semantic regularization-based mixup strategy to generate sufficient data to tackle the issue of imbalanced domain generalization. For metric learning method, Mo \textit{et al.} \cite{r57} considered instance-to-prototype distance, additional instance-to-instance and prototype-to-prototype distances and further proposed a distance-aware risk minimization framework through two novel losses. Despite the impressive performance of these methods, most of them predominantly explore spatial domain information while overlooking discriminative and generalizable information in the frequency domain. Our approach innovatively combines the extraction of domain-invariant features with domain augmentation. We leverage information in the Fourier space to reconstruct diverse frequency representations, resulting in a significant improvement on performance.

\subsection{Fourier-based Cross-Domain Research}
Early research \cite{r18} validated an important property of the Fourier transform: the phase components of the Fourier spectrum retain the high-level semantic information of the original signal, while the amplitude spectrum components contain low-level statistical information. Recently, in the researches of computer vision, Yang \textit{et al.} \cite{r23} proposed integrating the Fourier transform with DA by replacing partial amplitude spectra in the source domain images with those from the target domain, thereby reducing domain differences. Xu \textit{et al.} \cite{r24} introduced a Fourier-based data augmentation strategy that enables the model to capture phase information, achieving good generalization effects in unseen domains through linear interpolation of amplitude spectra which is similar to mix-up. Lin \textit{et al.} \cite{r41} uncovered that Deep Neural Networks have preference on some frequency components and used Deep Frequency Filtering (DFF) to explicitly extract the components in frequency domain of different transfer difficulties across domains in the latent space during training. The previous generalizable fault diagnosis method \cite{r42} only proposes a Fourier transform at the level of data pre-processing. To the best of our knowledge, Fourier-based data augmentation fault diagnosis is the firstly introduced in the cross-domain fault diagnosis. We propose a Fourier-based data augmentation reconstruction module and manifold deep metric learning to explore the potential information to improve the performance under unseen working conditions.

\begin{figure*}[ht]
\centerline{\includegraphics[width=\linewidth]{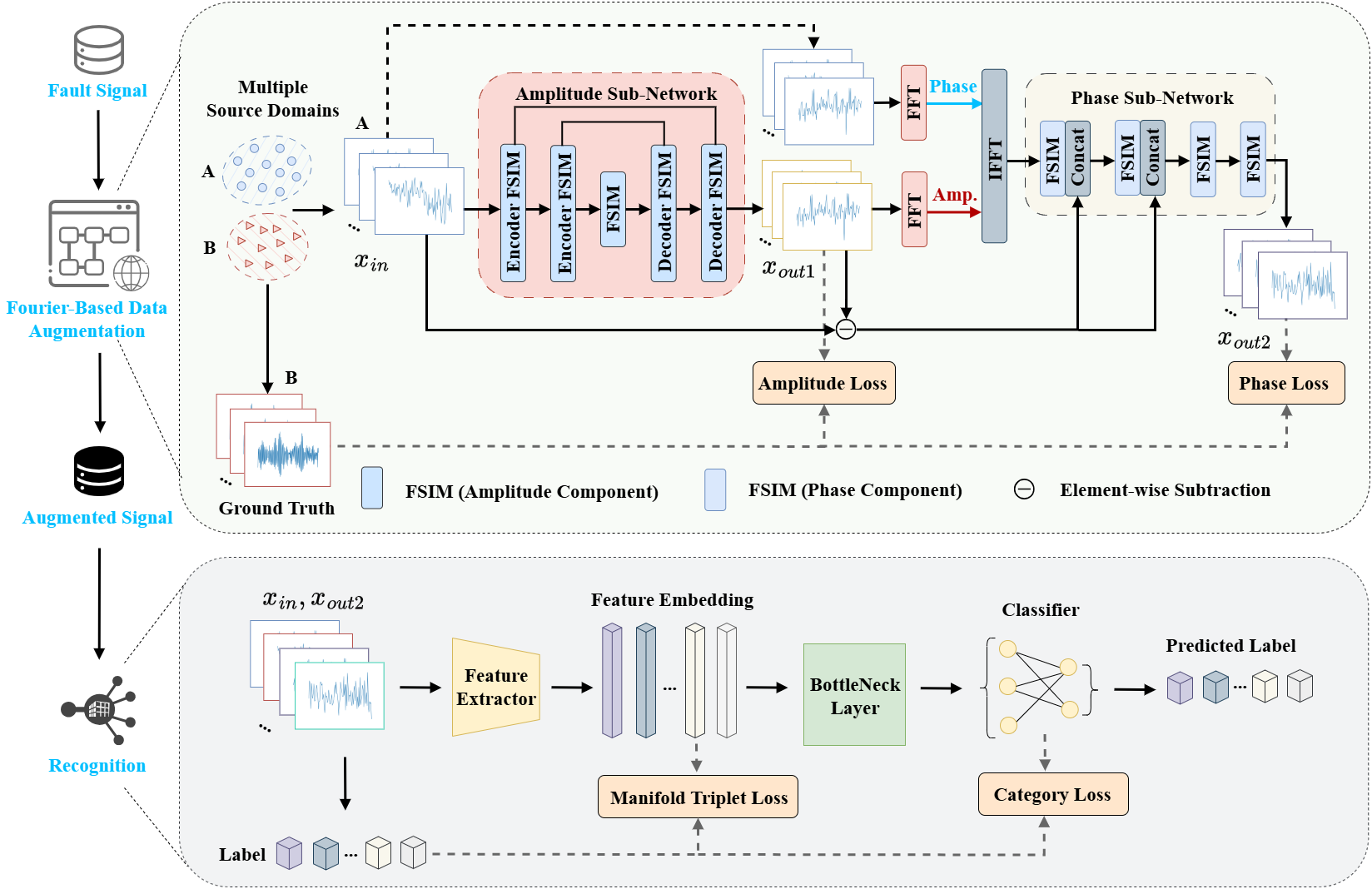}}\caption{The overall framework proposed in this paper consists of two main modules: the The Fourier-based Augmentation Reconstruction Network and the Recognition module. In the FARNet module, there are two sub-networks: an amplitude sub-network and a phase sub-network. The phase sub-network takes the fusion information of $\mathcal{F}^{-1}(\mathcal{A}(X_{out1}),\mathcal{P}(X_{in}))$ as input, guided by the residuals of the amplitude sub-network during the training process. Both components of the Frequency-Spatial Interaction Module (FSIM) serve as fundamental blocks in these networks, facilitating the feature extraction. In the final stage, both the original data and the augmented data are fed into the recognition network for fault diagnosis.}
\label{fig3}
\end{figure*}

\section{Method}
\subsection{Motivation and Background}
Influenced by industrial factors (i.e., working load, vibration frequency and temperature), the data distribution gap significantly affects the performance of fault diagnosis models. Previous works rarely focused on feature extraction in the frequency domain, despite its proven effectiveness in alleviating domain-shift issues. To address this gap, we introduce a Fourier-based Augmentation Reconstruction Network designed to effectively capture discriminative and generalizable information in the frequency domain by reconstructing frequency representations.

First, we revisit the definition and properties of the Fourier transform. For a signal $x$ with the shape of $C$$\times$$H$$\times$$W$, its Fourier transformation $\mathcal{F}(x)$ is formulated as:

\begin{equation}
\begin{aligned}
\mathcal{F}(x)(c,u,v) &= X(c,u,v)\\
& =\sum_{h=0}^{H-1} \sum_{w=0}^{W-1} x(c,h,w)e^{-j2\pi (\frac{h}{H}u+\frac{w}{W} v) },  \label{eq1} 
\end{aligned}
\end{equation}
where $C$, $H$, and $W$ are the channel, height, and width values of the signal, respectively.

\begin{figure*}[ht]
\centerline{\includegraphics[width=0.9\linewidth]{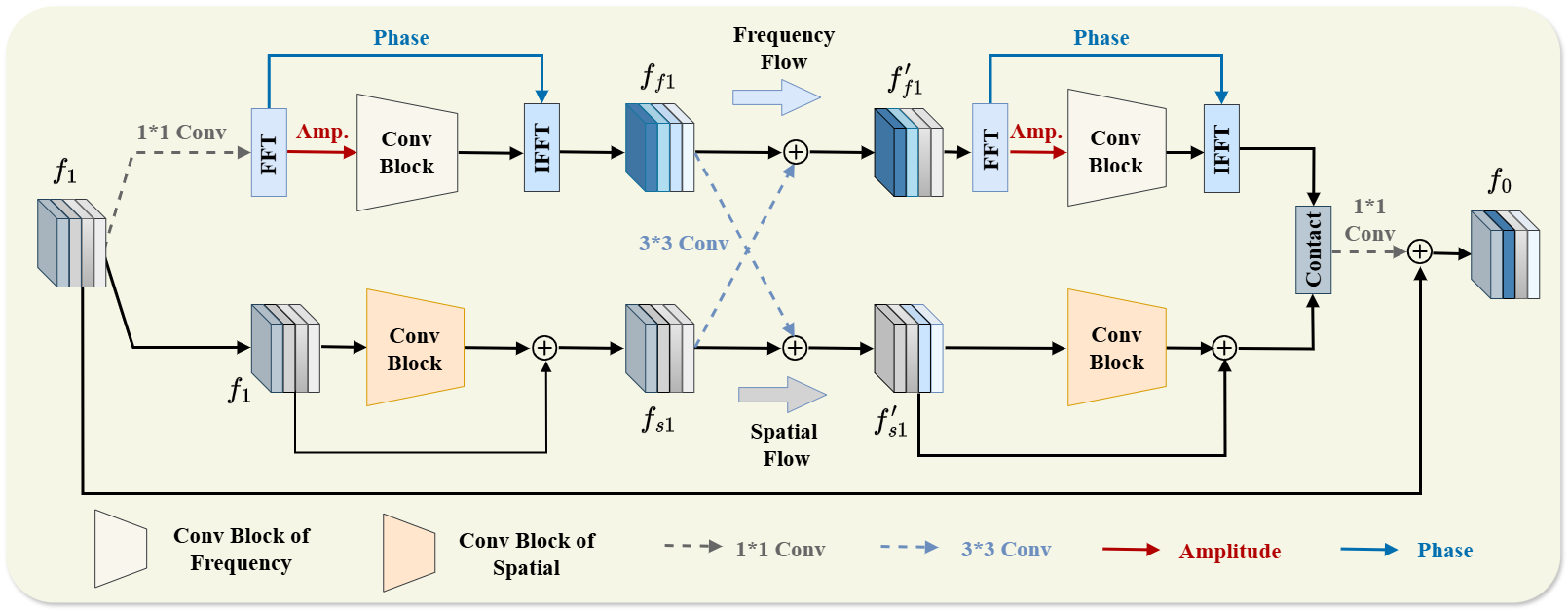}}\caption{The illustration depicts the amplitude format of the Frequency-Spatial Interaction Module (FSIM). The module comprises two flows: frequency flow and spatial flow. The frequency branch focuses on processing the amplitude component, bypassing the phase component to capture global representation. Simultaneously, the spatial branch utilizes residual blocks to acquire local feature representation.}
\label{fig4}
\end{figure*}

In the Fourier space, the amplitude and phase components are respectively expressed as:
\begin{equation}
\begin{aligned}
&\mathcal{A}(X(c,u,v))=\sqrt{R^{2}(X(c,u,v))+I^{2}(X(c,u,v))}, \\
&\mathcal{P}(X(c,u,v))=arctan\left [ \frac{I(X(c,u,v))}{R(X(c,u,v))}\right ], 
\label{eq2} 
\end{aligned}
\end{equation}
where $R(x)$ and $I(x)$ denote the real and imaginary parts of $X(c,u,v)$.

According to Fourier theory, the amplitude component $\mathcal{A}$ represents the stylistic information of the signal in the frequency domain and the phase component $\mathcal{P}$ represents the category information accordingly. As shown in Fig.\ref{fig1}, we observe distinct differences in the amplitude components of bearing fault data with the same category but different distributions, indicating significant variations in signal style. However, the phase components have strong coherent alignment which signifies identified signal categories. This implies that the phase components are associated with structural information, exhibiting less sensitivity to regional variations. According to this analysis, we enhance the diversity of signals from the source domain by progressively extracting and reconstructing both the amplitude and phase components. This approach reduces differences between domains, thereby improving the model's robustness.

\subsection{The Fourier-based Augmentation Reconstruction Network}
Based on the aforementioned analysis, we introduce a simple yet effective FARNet, illustrated in Fig.\ref{fig3}. The entire network comprises an amplitude spectrum sub-network and a phase spectrum sub-network, dedicated to reconstructing and enhancing amplitude and phase representations. Additionally, both sub-networks incorporate FSIM as a fundamental module to facilitate feature extraction and reconstruction by integrating both global and local spatial information. Further details on this aspect will be discussed in Section 3.3. The figure above illustrates the flow that samples from domain A are the input and the Fourier-based Augmentation Reconstruction Network strives to pull the input to align with the samples from domain B in the perspectives of amplitude and phase while . 

The amplitude sub-network adopts an encoder-decoder architecture, comprising five FSIMs for amplitude reconstruction, where the Encoder FSIMs are the squeeze modules, FSIM keeps the dimensions unchanged and the Decoder FSIMs strive to expand and reconstruct the $x_{in}$. Among multiple source domains, we select one domain (working condition) as the ground truth, denoted by $x_{gt}$, while the remaining source domain data are denoted as $x_{in}$ and the output of the amplitude sub-network is denoted as $x_{out1}$. They are represented in the Fourier space as $X_{gt}$, $X_{in}$, and $X_{out1}$. To guide the learning of amplitude representations, this sub-network is supervised by the ground truth amplitude $A(X_{gt})$. The loss function of the amplitude sub-network, denoted as $L_{amplitude}$, is described as:

\begin{equation}
\begin{aligned}
L_{amp} = \left \| \mathcal{A}(X_{out1})-\mathcal{A}(X_{gt}) \right \|_{1},  \label{eq3} 
\end{aligned}
\end{equation}
where $\left \| . \right \| _{1} $ represents the mean absolute error.

In the phase sub-network, four FSIMs are employed for the phase representation where the former two blocks fuse the residual from Amplitude Sub-Network. To ensure consistency of phase components, we use the reconstructed component $\mathcal{F}^{-1}(\mathcal{A}(X_{out1}),\mathcal{P}(X_{in}))$ instead of $x_{out1}$ as input to the sub-network. Additionally, considering variations in the sampling environment of bearing fault data, where structural components are sensitive to environmental factors, we leverage the residual of the outputs from amplitude sub-network and $x_{in}$ to capture domain shift differences (variations on speeds and fault diameters). By connecting the residuals of $x_{out1}$ and $x_{in}$ with the feature input of this sub-network through 1$\times$1 convolutions, we guide its learning process. The output of the phase sub-network is denoted as $x_{out2}$, and the loss function $L_{phase}$ for this sub-network is expressed as:

\begin{equation}
\begin{aligned}
L_{pha}= \left \| \mathcal{P}(X_{out2} )-\mathcal{P}(X_{gt} ) \right \|_{1}.  \label{eq4} 
\end{aligned}
\end{equation}

FARNet's data augmentation module is composed of these two sub-networks for end-to-end training. Therefore, the overall loss $L_{aug}$ of this augmentation module is formulated as:
\begin{equation}
L_{aug}=\lambda_{1} L_{amp}+\lambda_{2} L_{pha},   \label{eq6} 
\end{equation}
where $\lambda_{1}$ and $\lambda_{2}$ are the weight factors.

\subsection{Frequency-Spatial Interaction Module}
According to Fourier theory \cite{r25}, processing information in Frequency domain allows the model to capture global representations, while convolutional operations focus primarily on learning local spatial features. Motivated by this insight that enhances the learning of amplitude and phase representations, we introduce the Frequency-Space Interaction Module (FSIM) as the feature-fusion block for both the amplitude and phase sub-networks, enhancing their feature representation capabilities.

The amplitude format of FSIM is depicted in Fig.\ref{fig4}, comprising both a frequency flow and a spatial flow. Initially, $f_{1}$, representing features from the source domain data, is input into FSIM. In the spatial flow, it undergoes processing through a Conv Block, utilizing multiple $3\times3$ 2-D convolutional layers to enhance spatial domain information,  yielding $f_{s1}$. Simultaneously, the frequency flow processes $f_{1}$ with a $1\times1$ 2-D convolution, resulting in $f_{0}$. Subsequently, it reconstructs phase and amplitude in the frequency domain through Fourier transformation. Amplitude information is extracted using a Conv Block comprising multiple $1\times1$ 2-D convolutional layers, producing $f_{f1}$ as the output of frequency domain processing. The representation of $f_{f1}$ is defined as:

\begin{equation}
\begin{aligned}
f_{f_{1} }= \mathcal{F}^{-1} (Conv_{1\times1}(\mathcal{A}(F_{f_{0} } )),\mathcal{P}(F_{f_{0} } )),
\label{eq6} 
\end{aligned}
\end{equation}
where $Conv_{1\times1}$ denotes the $1\times1$ operation.


Then, $f_{f1}$ and $f_{s1}$ are separately fed into a $3\times3$ convolutional layer to facilitate the interaction of features between the frequency and spatial flows. In this case, $f_{f1}^{'}$ and $f_{s1}^{'}$ can be expressed as:

\begin{equation}
\begin{aligned}
&f_{f1}^{'} = f_{f1} +Conv_{3\times3} (f_{s1}),\\
&f_{s1}^{'} = f_{s1} +Conv_{3\times3} (f_{f1}),
\label{eq7} 
\end{aligned}
\end{equation}
where $Conv_{3\times3}$ is represented as the $3\times3$ operation.

As shown in Fig.\ref{fig4}, we observe that both $f_{f1}^{'}$ and $f_{s1}^{'}$ acquire complementary feature representations effectively aiding in the extraction of more discriminative features. Then, the frequency and spatial flows follow similar operations sequentially, culminating in an output $f_{0}$ from FSIM after passing through a final $1\times1$ convolutional layer.

Similarly, for the phase components of the FSIM, we replace the operations on the amplitude components in the frequency flow with operations on the phase components, while keeping the remaining structure unchanged.

\subsection{Manifold Triplet Loss}
In paper \cite{r39}, the author highlighted the effectiveness of using Euclidean neighbors as positive training instances to separate different classes into distinct subspaces. However, this may not be necessary in the manifold space. Therefore, we propose a modified triplet loss, namely the manifold triplet loss. In the manifold triplet loss, we replace the Euclidean distance in the triplet loss with the distance designed through a non-linear activation function $d_{new}(\cdot)$:

\begin{equation}
\begin{aligned}
d_{new}(x)=\left\{\begin{matrix}
 kx, & x>r\\
 \frac{x}{k},  & x\le r 
\end{matrix}\right.
\label{eq8} 
\end{aligned}
\end{equation}
where $x$ denotes the euclidean distance between two samples, $r$ is the constant threshold and $k$ is the weight factor, respectively.

By applying this activation operation to the Euclidean distances, we break the triangle inequality property of Euclidean space, where the sum of any two sides is greater than the third side. This characteristic enables us to explore not only the nearest neighbors but also encourages detour calculations when conducting depth measurements in manifold spaces. Specifically, we scale up long distances by a factor of $k$ while reducing short distances by a factor of $k$, encouraging detours when performing depth measurements in manifold spaces. Thus, the calculation of the manifold triplet loss is defined as follows:

\begin{equation}
\begin{aligned}
&\
L_{manifold-triplet}=max(d_{new}^{p}(x) - d_{new}^{n}(x) + \gamma, 0), \label{eq9}
&
\end{aligned}
\end{equation}
where $d_{new}^{p}(x)$ and $d_{new}^{n}(x)$ denote the manifold distance of the hardest positive sample and negative sample in a batch respectively, and $\gamma$ represents the margin of triplet loss.

Besides, the fault feature vectors $V$ are extracted from the original signals $x_{in}$ and signals  $x_{out2}$ generated by the Fourier-based Augmentation Reconstruction Module through ResNet18 and used for fault recognition, the predicted probability $p(v_{i})$ is computed by the softmax function as follows:

\begin{equation}
\begin{aligned}
&\
p(v_{i}) = \frac{e^{v_{i}}}{ {\textstyle \sum_{N_{cls}}^{j=1}}e^{x_{j}} }, i=1,2,...,N_{cls} \label{eq10}
&
\end{aligned}
\end{equation}
where $v_{i}$ denotes the i-th bearing fault instance, $N_{cls}$ is the number of fault categories. The greatest probability is defined as the final classification result. Based on the predicted results and ground-truth labels, the recognition module is optimized with the cross-entropy loss as below:

\begin{equation}
\begin{aligned}
&\
L_{clf} = - \sum_{N_{cls}}^{i=1} log(p(v_{i})), \label{eq11}
&
\end{aligned}
\end{equation}
where $y_{i}$ is the ground-truth labels of the i-th bearing fault instance.

\subsection{Optimization}
Aggregating all loss functions, we can obtain the training objective of FARNet as follows:

\begin{equation}
\begin{aligned}
L_{total}= L_{aug}+L_{clf}+\alpha L_{manifold-triplet},   
\label{eq12} 
\end{aligned}
\end{equation}
where $\alpha$ is the weight factor to balance the convergence of the two modules.

\section{Experiment}
In this section, we demonstrate the superiority of our approach compared to several Domain Adaption and Domain Generalization methods. We also conduct ablation experiments on some hyperparameters and finally validate the effectiveness of our approach through visualization.

\subsection{Datasets and Settings}
\begin{figure}[!t]
\centerline{\includegraphics[width=\linewidth]{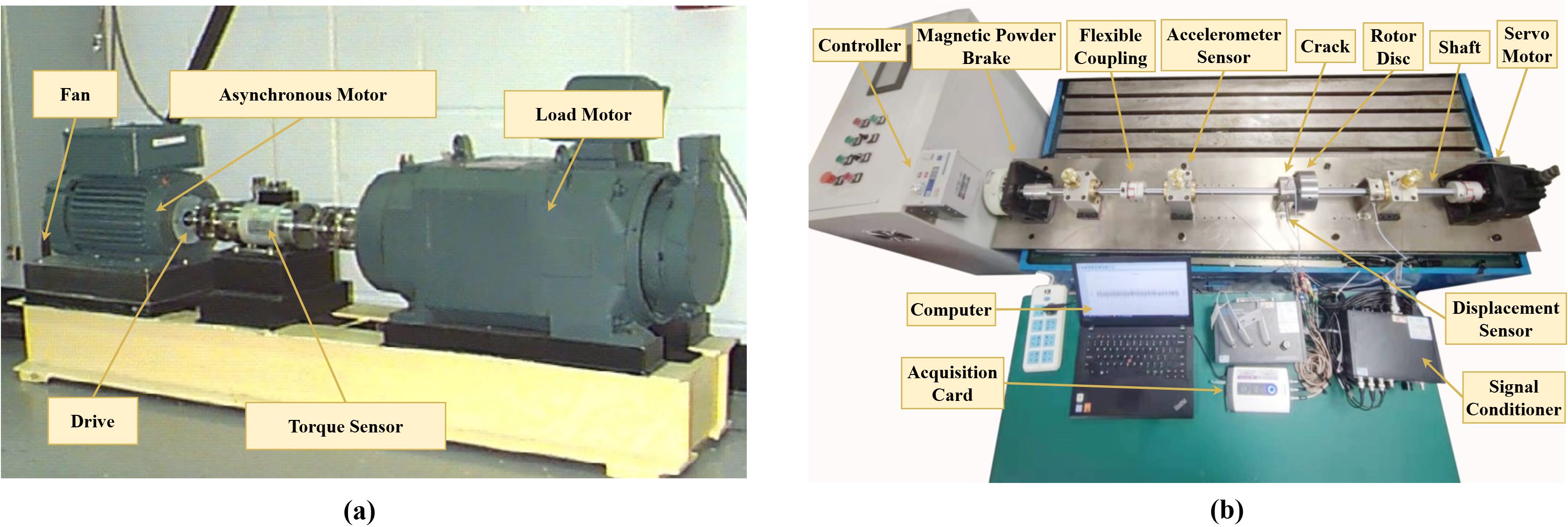}}\caption{(a) CWRU experimental setup for machinery fault diagnosis \cite{r32}, (b) SJTU experimental setup for machinery fault diagnosis.}
\label{fig5}
\end{figure}

\textbf{Datasets.} We validated the effectiveness of our model on both a public dataset and a dataset collected by ourselves. The details of these datasets are as follows: 

(1) \textbf{CWRU} \cite{r32}: The bearing dataset was collected from the rolling bearing test bench at Case Western Reserve University. As shown in Fig.\ref{fig5}(a), the entire rolling bearing test bench mainly consists of the fan end, drive end, torque sensor, asynchronous motor, and load motor. Four types of loads were simulated in total: 0 hp, 1 hp, 2 hp, and 3 hp. There are four statutes for the rolling bearings: Normal (N), Inner Race Fault (IRF), Ball Fault (BF), and Outer Race Fault (ORF). The raw vibration signals collected from the four states are sampled at a frequency of 12 kHz and each fault type has three fault sizes (7, 14, 21 mils). In this experiment, we selected fault diagnosis data from the drive end under a 3hp load, where each fault type contains 200 training samples and 100 test samples. The shape of an individual training sample is 1$\times$3600$\times$1. This is used to verify the effectiveness of FARNet in multi-source domain generalization (including CW$_{7\_14}$$\rightarrow$CW$_{21}$, CW$_{7\_21}$$\rightarrow$CW$_{14}$ and CW$_{14\_21}$$\rightarrow$CW$_{7}$). 

\begin{figure}[!t]
\centerline{\includegraphics[width=\linewidth]{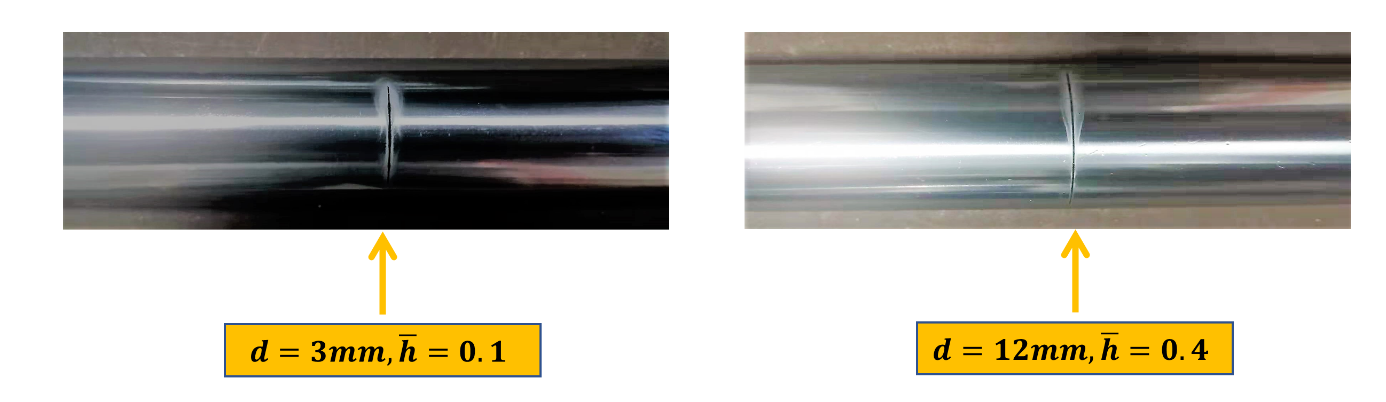}}\caption{Picture of rotor crack in SJTU dataset.}
\label{fig6}
\end{figure}

(2) \textbf{SJTU}: The cracked rotor dataset was collected on our test bench, as shown in Fig.\ref{fig5} (b). The rotor test bench mainly consists of a controller, magnetic powder brake, accelerometer sensor, flexible couplings, rotor disc, shaft, and servo motor. While the rotor signal acquisition system consists of a computer, acquisition card, signal conditioner, and displacement sensors mounted on the rotor test bench. A transverse crack was machined at the shaft's midpoint to simulate a breathing crack occurring during rotor operation. The experiment collected rotor operational data at speeds of 1000, 1200, 1400, 1600, and 1800 r/min with a sampling frequency of 6400Hz for different crack depths. In order to collect the vibration data of the rotor when there is a crack fault, a transverse crack is line-cut at the middle position of the rotor shaft to simulate the breathing crack produced during the actual operation of the rotor. The crack location is located next to the disc, and the crack form is shown in Fig.\ref{fig6}. Two types of crack depth schematic are shown in Fig.\ref{fig6}, the crack depth is 3mm and 12mm respectively, the rotor diameter is 30mm, the corresponding relative crack depth is 0.1 and 0.4 respectively. The crack depths were set at 0 mm, 3 mm, 6 mm, 9 mm, 12 mm, 15 mm, and 18 mm. Each crack depth serves as a category. In this experiment, we use data from seven categories at three speeds of 1000, 1200, and 1400 r/min (denoted as SJ$_{1}$, SJ$_{2}$, SJ$_{3}$) to validate the effectiveness of multi-source domain generalization (including SJ$_{1\_2}$$\rightarrow$SJ$_{3}$, SJ$_{1\_3}$$\rightarrow$SJ$_{2}$ and SJ$_{2\_3}$$\rightarrow$SJ$_{1}$). Each fault category includes 800 training samples and 200 test samples. The size of an individual training sample is 6$\times$2048$\times$1, this data is obtained from 6 channels.

\begin{table}
\small
\centering
\caption{Details of the health states for two fault datasets}
\label{Details of the health states for two fault datasets}
\begin{tabular}{cccccc} 
\hline \hline
\multirow{2}{*}{Dataset} & \multicolumn{3}{c}{Working Condition  }                                         & \multirow{2}{*}{health types}  & \multirow{2}{*}{No.of samples}  \\ 
\cline{2-4}
                         & Code           & Speed                    & Fault diameter                      &                                &                                 \\ 
\hline
\multirow{3}{*}{CWRU}    & CW$_{7}$  & \multirow{3}{*}{1730rpm} & 7mil                                & \multirow{3}{*}{N/IR/B/OR}     & \multirow{3}{*}{4$\times$300}          \\
                         & CW$_{14}$ &                          & 14mil                               &                                &                                 \\
                         &  CW$_{21}$ &                          & 21mil                               &                                &                                 \\ 
\hline
\multirow{3}{*}{SJTU}    &  SJ$_{1}$  & 1000rpm                  & \multirow{3}{*}{0/3/6/9/12/15/18mm} & \multirow{3}{*}{0/1/2/3/4/5/6} & \multirow{3}{*}{7$\times$1000}          \\
                         &  SJ$_{2}$  & 1200rpm                  &                                     &                                &                                 \\
                         &  SJ$_{3}$  & 1400rpm                  &                                     &                                &                                 \\
\hline \hline
\end{tabular}
\end{table}

\subsection{Implementation Details}
The method we proposed is implemented using the PyTorch framework on Python 3.9 and a single NVIDIA 2080Ti GPU. During the training process, the framework is trained for 50 epochs with a batch size of 128. And SGD optimizer is employed to update the overall framework with momentum of 0.9. Due to the different sensitivity of the Fourier-based data augmentation network and the Recognition network, the initial learning rates are set to 0.001 and 0.01, respectively. To train the manifold triplet loss, each batch is constructed by PK sampler which comprises four types where each type is composed of 32 instances in a batch. The constant threshold $r$ is calculated by the mean distance in a batch and the margin is 0.3 in manifold triplet loss. The value of $\alpha$ is set to 0.01 in the training procedure. The other hyper-parameters of losses are illustrated in the section 4.4.2 and 4.4.3. To ensure the reliability and effectiveness of the experimental results, we conducted experiments for each method five times for each task.

\begin{table*}
\renewcommand\arraystretch{1.7}
\centering
\caption{Comparison on the performance of multi-source domain generalization methods}
\label{table2}
\resizebox{15cm}{!}{
\begin{tabular}{c|cccc|cccc}
\hline\hline
\multirow{2}{*}{Method} & \multicolumn{4}{c|}{CWRU}                                                                   & \multicolumn{4}{c}{SJTU}                                                                   \\ \cline{2-9} 
                        & CW$_{7\_14}$$\rightarrow$CW$_{21}$                     & CW$_{7\_21}$$\rightarrow$CW$_{14}$                   & CW$_{14\_21}$$\rightarrow$CW$_{7}$                     & Avg.                  & SJ$_{1\_2}$$\rightarrow$SJ$_{3}$                    & SJ$_{1\_3}$$\rightarrow$SJ$_{2}$                   & SJ$_{2\_3}$$\rightarrow$SJ$_{1}$                     & Avg.                  \\ \hline
CNN$\_$C \cite{r63}               & 50.00$\pm$4.91             & 25.00$\pm$6.52          & 25.75$\pm$5.76             & 33.58$\pm$7.00          & 40.00$\pm$6.73          & 44.14$\pm$2.13        & 49.71$\pm$4.17          & 44.62$\pm$4.34          \\
ResNet18 \cite{r64}                & 71.50$\pm$4.91             & 37.20$\pm$7.52          & 65.10$\pm$8.76             & 57.93$\pm$7.06          & 51.91$\pm$6.73          & 59.11$\pm$2.13        & 55.00$\pm$4.17          & 55.34$\pm$4.34          \\
MMD \cite{r8}                     & 74.40$\pm$0.34            & 44.50$\pm$3.16        & 65.65$\pm$1.15           & 61.52$\pm$1.55          & 70.94$\pm$6.89         & 72.12$\pm$4.16         & 77.85$\pm$5.59           & 73.64$\pm$5.55          \\
SNR \cite{r28}                     & 68.15$\pm$9.24           & 44.85$\pm$7.77         & 66.51$\pm$8.94           & 59.84$\pm$8.65           & 57.49$\pm$6.56         & 65.69$\pm$6.49        & 56.06$\pm$3.09          & 59.75$\pm$4.24          \\
Mixup \cite{r26}                  & 65.50$\pm$0.94            & 49.80$\pm$2.61          & 78.75$\pm$4.01           & 64.68$\pm$2.52           & 55.35$\pm$3.93          & 58.97$\pm$4.22         & 60.80$\pm$6.49          & 58.37$\pm$4.88          \\
MixStyle \cite{r27}               & 73.05$\pm$1.67           & 47.65$\pm$5.81          & 75.30$\pm$4.61            & 65.33$\pm$4.03              & 56.74$\pm$3.94         & 66.97$\pm$5.01         & 64.29$\pm$4.91          & 62.67$\pm$4.62          \\
DARM \cite{r55}                  & 54.50$\pm$9.94           & 44.55$\pm$2.29          & 64.30$\pm$7.83            & 54.45$\pm$6.69              & 47.17$\pm$1.89         & 60.86$\pm$2.27         & 47.65$\pm$3.63          & 51.89$\pm$2.60          \\
IEDGNet \cite{r58}                  & 62.90$\pm$8.77           & 47.35$\pm$9.80          & 73.10$\pm$7.30            & 61.12$\pm$8.62              & 54.17$\pm$3.43         & 49.31$\pm$6.90         & 53.91$\pm$4.52          & 52.46$\pm$4.95          \\
CCDG \cite{r57}             & 62.42$\pm$7.51           & 48.22$\pm$3.44          & 84.42$\pm$10.14            & 64.69$\pm$7.03              & 46.54$\pm$5.71         & 64.35$\pm$2.42         & 56.60$\pm$5.08          & 55.83$\pm$4.40          \\ \hline
\textbf{FARNet(ours)}   & \textbf{99.24$\pm$0.56} & \textbf{53.13$\pm$1.70}          & \textbf{99.67$\pm$0.67} & \textbf{84.01$\pm$0.98} & \textbf{82.24$\pm$2.45} & \textbf{80.17$\pm$1.91} & \textbf{84.28$\pm$1.30} & \textbf{82.23$\pm$1.92} \\ \hline\hline
\end{tabular}}
\end{table*}

\begin{figure}[!t]
\centerline{\includegraphics[width=\linewidth]{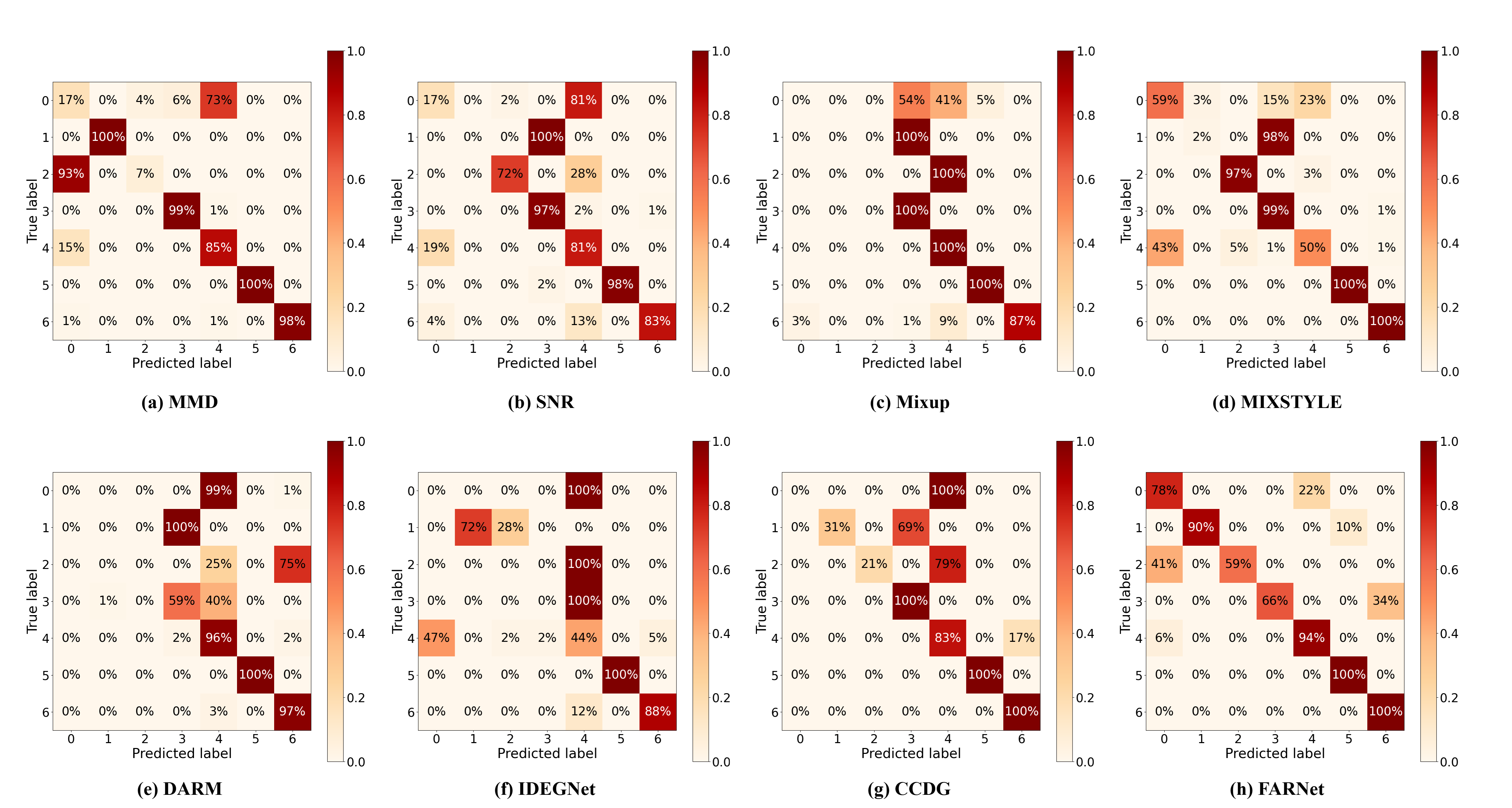}}\caption{Confusion matrices of different diagnostic methods under SJ$_{1\_2}$$\rightarrow$SJ$_{3}$ condition of SJTU dataset.The first line is the classical domain adaptation, domain generalization approach and the second line is a domain generalization approach for bearing fault diagnosis.}
\label{fig7}
\end{figure}

\subsection{Comparative Experiment}
In this experiment, we compared our proposed method with three categories of benchmark models: classical DA methods MMD \cite{r8}, well-known DG methods (including SNR \cite{r28}, Mixup \cite{r26}, MixStyle \cite{r27}) and the state-of-the-art DG methods for fault diagnosis (including CNN$\_$C \cite{r63}, DARM \cite{r55}, IEDGNet \cite{r58} and CCDG \cite{r57}) in the setting of multi-source domain generalization. For the classical DA, well-known DG methods and DARM, the feature extractors are set to ResNet18 \cite{r64} (Baseline). And the backbone of CNN$\_$C is a four-layer convolutional neural network. The relevant domain generalization experiments are conducted based on the GitHub project\footnote{\url{https://github.com/CHAOZHAO-1/Domain-generalization-fault-diagnosis-benchmark}}. The average accuracies and their corresponding standard deviations of all methods are illustrated in Table \ref{table2}.

It can be observed from Table \ref{table2} that the final average accuracy of our FARNet method reaches the highest average accuracy of 84.01$\%$ on the CWRU dataset and 82.32$\%$ on the SJTU dataset, which are superior to the comparison methods. For the most commonly used DA-based diagnostic methods, the average diagnostic accuracies of MMD is 61.52$\%$ on the CWRU dataset and 73.64$\%$ on the SJTU dataset, respectively, which fall significantly behind our method. The main reason is that MDD only strives to narrow the distribution gap of the source domain through explicit measurement or adversarial training, making them highly susceptible to noise in the data, resulting in the degradation of model's robustness. We also compare conventional DG methods (SNR, Mixup and MixStyle) that aim to enhance the generalizable ability through the extraction of domain-invariant features and domain augmentation. However, the average accuracies of them on CWRU are 59.84$\%$, 64.68$\%$ and 65.33$\%$ while 59.75$\%$, 58.37$\%$ and 62.67$\%$, which are lower than our method. The main reason is that they only focus on one aspect of the two methods, thus they exhibit limitations in generalization performance. For cross-domain fault diagnosis methods, they merely consider the improvement on the deep metric learning and regularization which neglect the benefit of data augmentation and achieve few enhancement compared to baseline model. 

Fourier-based data augmentation and deep metric learning are introduced in our proposed method. This approach empowers the model with robust learning of invariant diagnostic knowledge for data augmentation, making it better suited for fault diagnosis. Consequently, its performance on the CWRU dataset surpasses that of conventional DG methods and the state-of-the-art generalizable fault diagnosis methods and achieves superior and robust results with $84.01\%$ accuracy and $0.98\%$ standard deviation (The highest accuracy is $99.67\%$). However, the model's accuracy significantly decreased as it conducted on the more complicate SJTU dataset where the working conditions are differed in speed. The highest average accuracy of our approach in the scenario of SJTU is only $84.28\%$, but our method is still the most stable one. Besides, we also obtain the confusion matrices of the representative fault diagnosis method in Fig.\ref{fig7} which demonstrate that FARNet demonstrates superior performance in accurately classifying instances of Class 0 and Class 1, whereas other methods encounter significant challenges in achieving comparable recognition accuracy for the two classes.

\subsection{Ablation Experiment}
To further verify the significance of the key components of the proposed FARNet, a series of ablation experiments were conducted. For a fair comparison, the parameters are consistent with those of the algorithm proposed in this paper.

\begin{table*}
\renewcommand\arraystretch{1.7}
\centering
\caption{The performance of key components on the CWRU and SJTU dataset.}
\label{table3}
\resizebox{15cm}{!}{
\small
\begin{tabular}{c|cccc|cccc}
\hline\hline
\multirow{2}{*}{Method} & \multicolumn{4}{c|}{CWRU}                                                                   & \multicolumn{4}{c}{SJTU}                                                                   \\ \cline{2-9} 
                        & CW$_{7\_14}$$\rightarrow$CW$_{21}$                     & CW$_{7\_21}$$\rightarrow$CW$_{14}$                   & CW$_{14\_21}$$\rightarrow$CW$_{7}$                     & Avg.                  & SJ$_{1\_2}$$\rightarrow$SJ$_{3}$                    & SJ$_{1\_3}$$\rightarrow$SJ$_{2}$                   & SJ$_{2\_3}$$\rightarrow$SJ$_{1}$                     & Avg.                  \\ \hline
M1               & 71.50$\pm$4.91             & 37.20$\pm$7.52          & 65.10$\pm$8.76             & 57.93$\pm$7.06          & 51.91$\pm$6.73          & 59.11$\pm$2.13        & 55.00$\pm$4.17          & 55.34$\pm$4.34          \\
M2                    & 90.18$\pm$3.99           & 44.96$\pm$4.04         & 93.95$\pm$4.53           & 76.36$\pm$4.19           & 80.17$\pm$5.31         & 76.86$\pm$6.16        & 83.71$\pm$4.46          & 80.25$\pm$5.25          \\
M3                     & 93.20$\pm$1.47           & 48.26$\pm$1.74        & 98.99$\pm$1.20           & 80.15$\pm$1.47          & 81.31$\pm$4.40         & 78.93$\pm$6.04         & 83.80$\pm$2.87           & 81.34$\pm$4.43          \\
\textbf{M4}   & \textbf{99.24$\pm$0.56} & \textbf{53.13$\pm$1.70}          & \textbf{99.67$\pm$0.67} & \textbf{84.01$\pm$0.98} & \textbf{82.24$\pm$2.45} & \textbf{80.17$\pm$1.91} & \textbf{84.28$\pm$1.30} & \textbf{82.23$\pm$1.92} \\ \hline\hline
\end{tabular}}
\end{table*}

\begin{figure}[ht]
\centerline{\includegraphics[width=\linewidth]{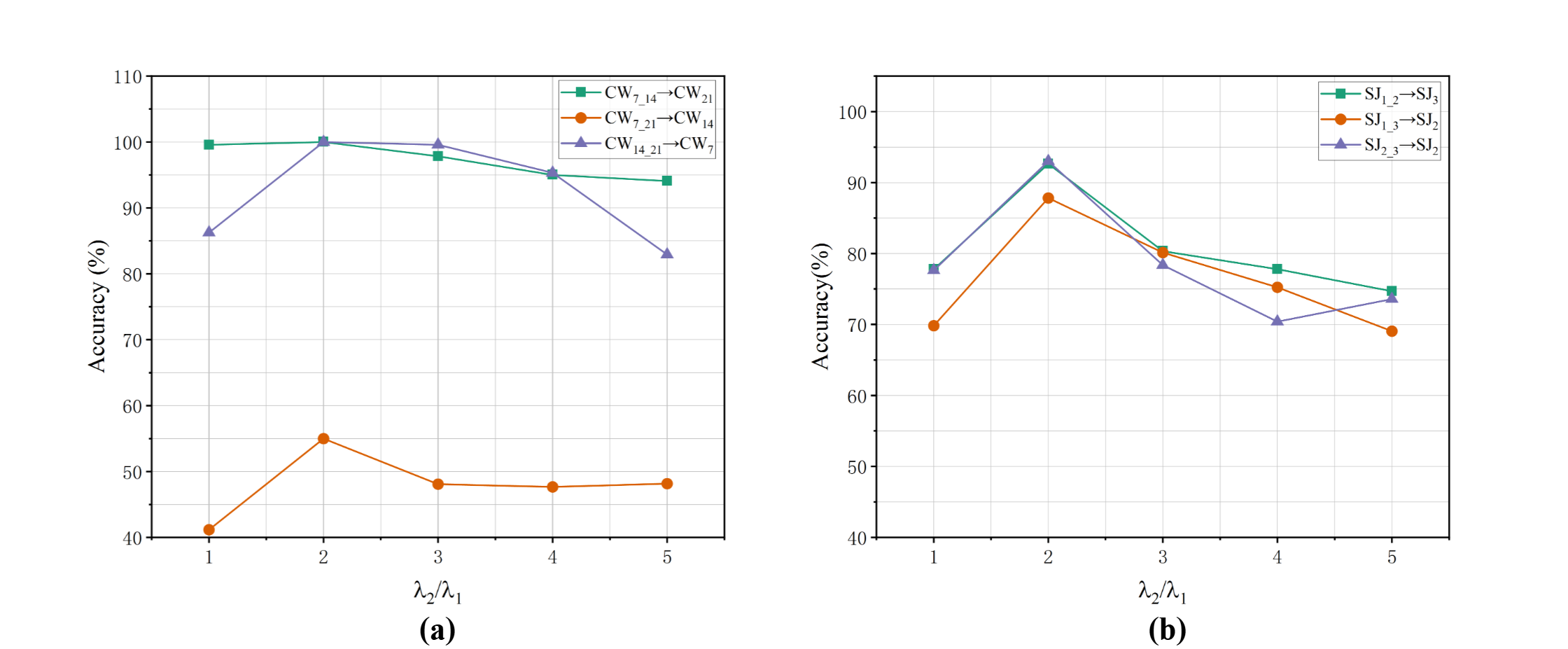}}\caption{The ablation experiments of hyperparameter $\lambda_{2}/\lambda_{1}$ for $L_{aug}$ on two datasets ((a) is CWRU and (b) is SJTU).}
\label{fig8}
\end{figure}

\begin{figure}[ht]
\centerline{\includegraphics[width=\linewidth]{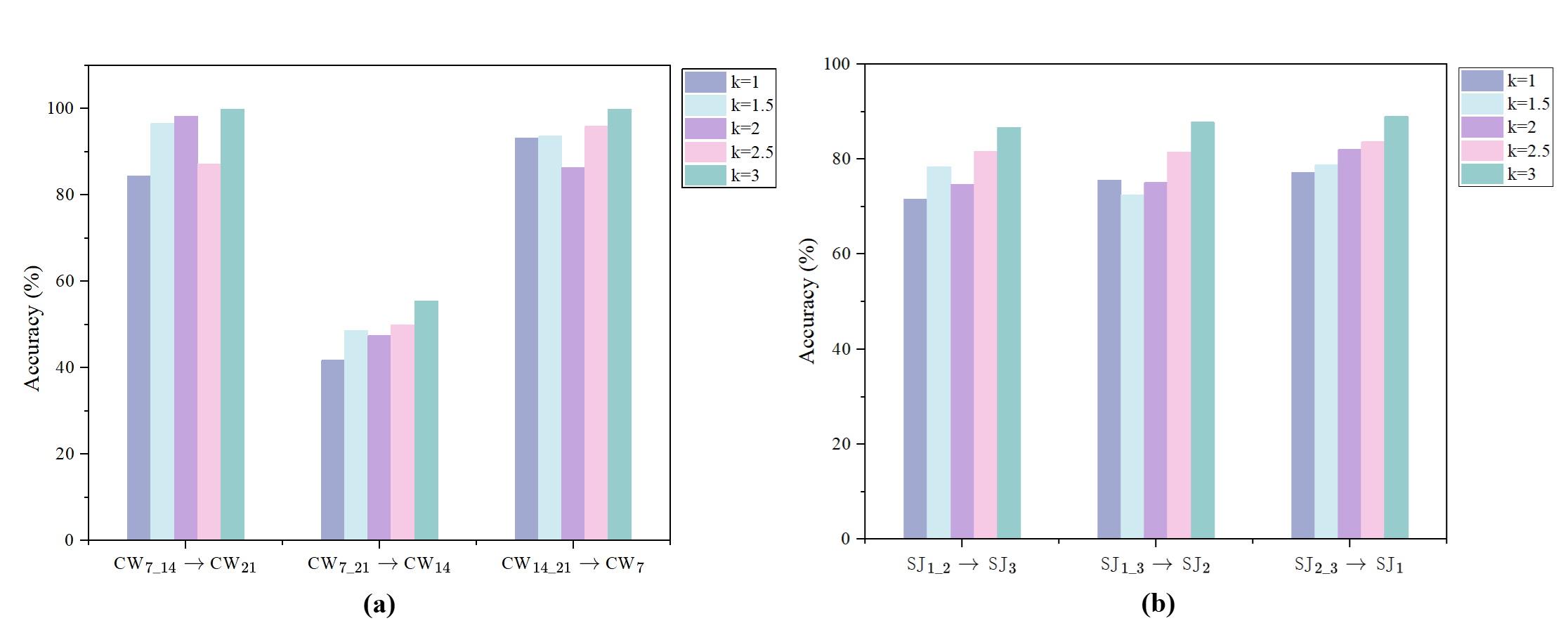}}\caption{The ablation experiments of hyperparameter k for $L_{manifold-triplet}$ on two datasets ((a) is CWRU and (b) is SJTU).}
\label{fig9}
\end{figure}

\begin{figure}[ht]
\centerline{\includegraphics[width=\linewidth]{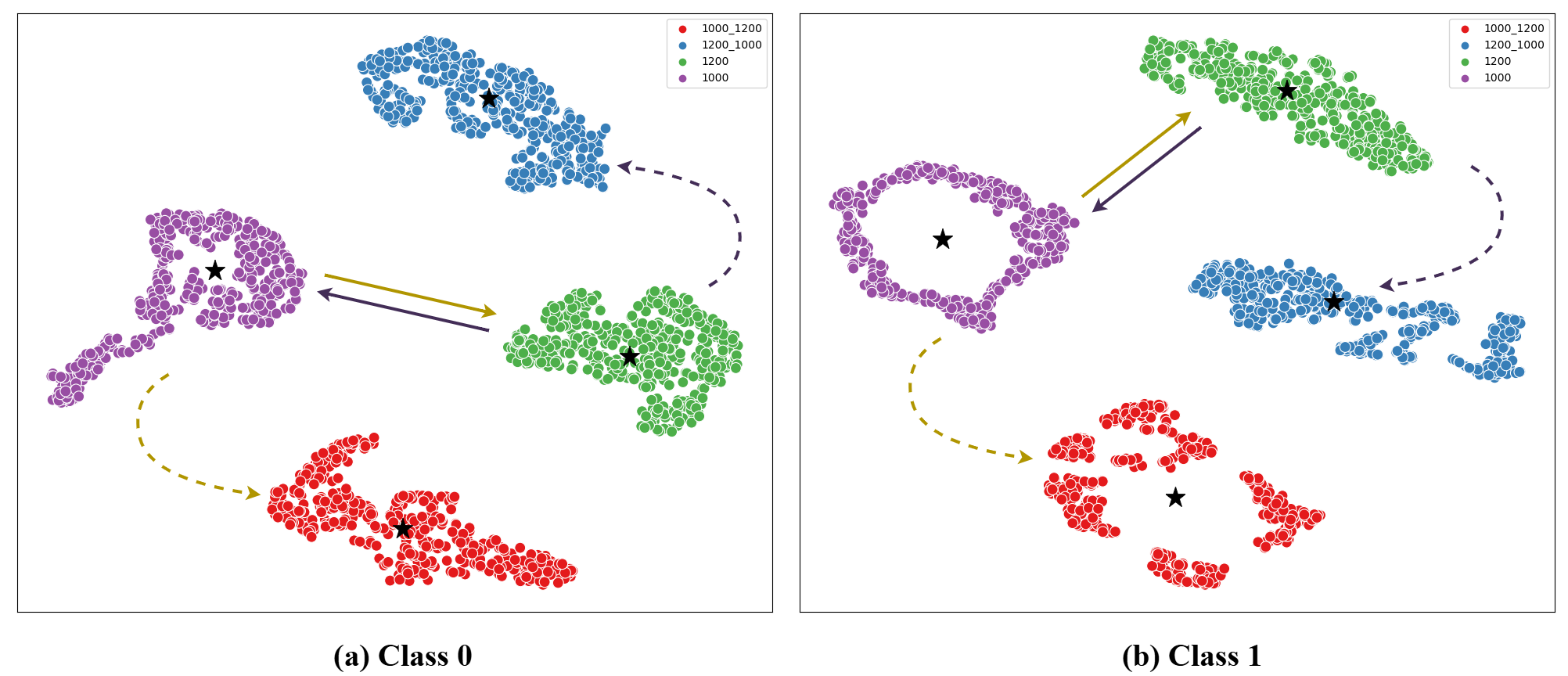}}\caption{The T-SNE results of first and second categories from the four domains (1000 and 1200 are real domains, while 1000$\_$1200 and 1200$\_$1000 are synthetic domains generated by data augmentation) on the SJTU dataset. Five-angle stars are the centers for the four domain clusters, and solid lines represent the alignment targets while dashed lines depict the actual generation process.
}
\label{fig10}
\end{figure}

\begin{figure}[ht]
\centerline{\includegraphics[width=\linewidth]{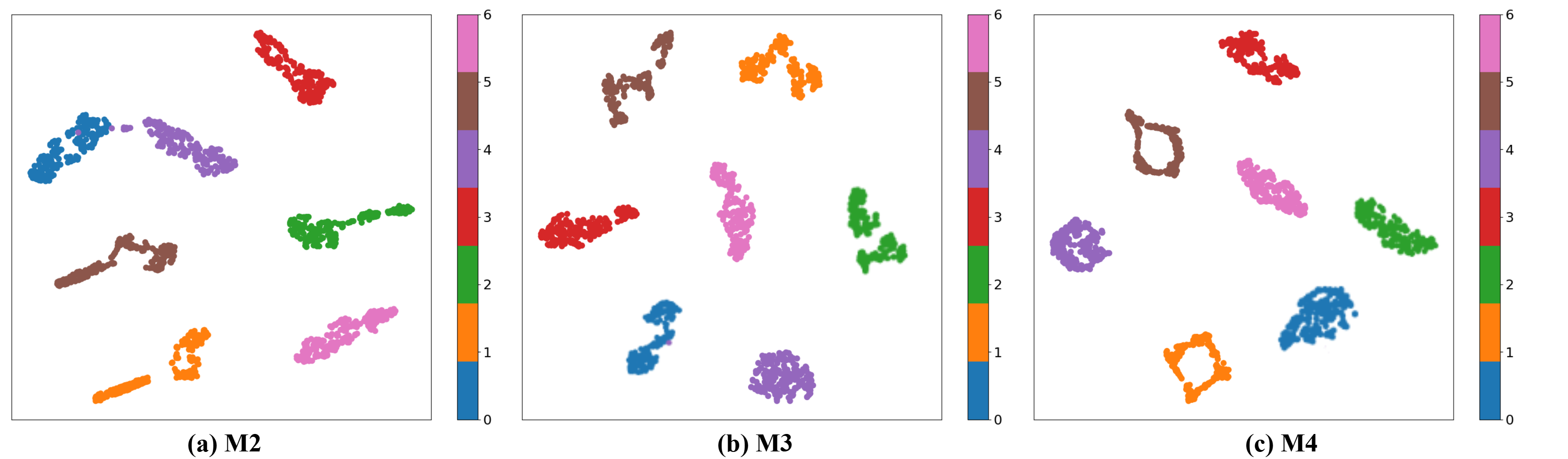}}\caption{T-SNE of M2, M3, and M4 diagnostic methods under SJ$_{1\_2}$$\rightarrow$SJ$_{3}$ condition of SJTU dataset.}
\label{fig11}
\end{figure}

\subsubsection{Ablation on the key components}
We sequentially validated the performance of each key component on the two benchmarks, as shown in Table \ref{table3}. Here, M1 represents the baseline (ResNet18), M2 represents only the augmentation module, M3 represents the combination of the augmentation module and triplet loss, and M4 refers to the FARNet model proposed in this paper. The results clearly indicate that FARNet achieved the highest diagnostic accuracy in the series. Examining the diagnostic results of M1 and M2, we observe that the augmentation module can boost accuracy by up to 28.85\% and 28.71\% on the two datasets, respectively. Comparing modules M3 and M4, we find that manifold triplet loss effectively aggregates data features, reduces inter-class distances, and enhances generalization.

We sequentially validate the performance of each key component on the two benchmarks, as shown in Table \ref{table3}. Where M1 represents the baseline, which is ResNet18, M2 represents only the augmentation module, M3 represents the combination of the augmentation module and triplet loss, M4 represents the combination of the augmentation module and manifold triplet loss, which refers to the FARNet model proposed in this paper. It clearly shows that FARNet brings the highest diagnostic accuracy in the series. From the diagnostic results of M1 and M2, we find that the augmentation module can boost the performance up to 24.43\% and 17.67\% on the two datasets, respectively. Comparing modules M3 and M4, we can observe that manifold triplet loss can effectively aggregate data features, reduce inter-class distances, and enhance generalization. 

\subsubsection{Hyperparameter Ablation in $L_{aug}$}
To explore the impact of hyperparameters in the augmentation network on model performance, we conducted an analysis focusing on weight factors $\lambda_{1}$ and $\lambda_{2}$. We considered candidate values for $\lambda_{2}/\lambda_{1}$, including 1, 2, 3, 4, and 5. Through multiple experiments, the results, as depicted in Fig.\ref{fig8}, reveal that our model achieves the optimal fault diagnosis accuracy on both datasets when the ratio of $\lambda_{2}$/$\lambda_{1}$ is set to 2. Besides, we set $\lambda_{1}$ is 0.1 and $\lambda_{2}$ is 0.2, respectively.

To investigate the effectiveness of hyperparameters in the augmentation network affecting model performance, we conducted a hyperparameter analysis on weight factors $\lambda_{1}$ and $\lambda_{2}$. The candidate values of $\lambda_{2}/\lambda_{1}$ are 1, 2, 3, 4, 5. Through multiple experiments, we obtained the results as shown in Fig.\ref{fig8}. From the figure, it can be observed that our model achieves the best fault diagnosis accuracy on both datasets when $\lambda_{2}/\lambda_{1}$ = 2.

\subsubsection{Ablation study on $L_{manifold-triplet}$}
\textbf{The sensitivity of $k$:} To identify the optimal feature aggregation model for the manifold triplet loss, we conducted ablation experiments by varying the values of $k$. Specifically, we set $k$ to 1, 1.5, 2, 2.5, and 3 in this experiment. As depicted in Fig.\ref{fig9}, the model achieves the best performance when $k$ is set to 3. This result suggests that, at $k=3$, the data features exhibit clearer boundaries, smaller intra-class distances, and larger inter-class distances, facilitating more effective fault diagnosis. To explore the optimal feature aggregation model of manifold triplet loss, we conducted ablation experiments on $k$ values. In this experiment, we set the $k$ values to 1, 1.5, 2, 2.5, and 3. From Fig.\ref{fig9}, we can observe that the model performs best when $k$ is set to 3. This indicates that when k is set to 3, the data features exhibit more distinct boundaries, reduced intra-class distances, and increased inter-class distances, thereby facilitating more accurate fault diagnosis.

\subsubsection{Visualization} 

To further verify the performance of our proposed methods, two visualization experiments are employed in this section on the diversity of the synthetic fault domains and the effectiveness of the manifold triplet loss in Fig.\ref{fig10} and Fig.\ref{fig11}.

\textbf{Visualization on the synthetic fault domains:} The synthetic fault domains are generated through multi-source domain generalization, and we utilize the T-SNE to visualize the output $x_{out2}$ on the SJTU dataset, which verifies the diversity of the synthetic fault domains in the Fig.\ref{fig10}. Two categories from working conditions (speed 1000 and 1200) are employed to obtain the two synthetic fault domains through the Fourier-based Augmentation Reconstruction Network. It is evident that this domain augmentation method can effectively enrich the diversity of fault domains while samples from different domains are mutually irrelevant.

\textbf{Visualization on different components:} For the visualization of manifold triplet loss, we list the three methods (M2, M3 and M4) conducted on the scenario of SJ$_{1\_2}$$\rightarrow$SJ$_{3}$ with T-SNE on the SJTU dataset in the Fig.\ref{fig11}. Fourier-based data augmentation is introduced in M2, but the intra-class distributions are relatively dispersed. Introducing triplet loss has made some classes more compact, but there still exists the issue that some classes are divided into multiple clusters. As depicted in Fig.\ref{fig11} (c), manifold triplet loss strives to refine the decision boundary and we obtain a more narrow intra-class data distribution which gathers each category into a circle.

\section{Conclusion}
This paper introduces FARNet, a Fourier-based Data Augmentation Reconstruction Network, designed to excel in fault diagnosis domain generalization. FARNet employs a data reconstruction enhancement strategy to alleviate domain shifts among multi-source domain data, facilitating the exploration of domain-invariant information generalization to previously unseen domain data. The FSIM is integrated into both the amplitude and phase spectrum sub-network of FARNet, enhancing feature extraction by addressing both global information and local spatial features. Additionally, the usage of manifold triplet loss ensures feature clustering in the manifold space. The effectiveness of FARNet is validated on the CWRU and SJTU datasets. The method is firstly introduced to the bearing fault diagnosis datasets, and we hope our method can be applied to various industrial scenarios in the future. Despite the promising results on the multi-source cross unknown working conditions, it's still a potential challenge for the model to recognize the samples from unknown fault categories. In future research, we'll explore the possibility to extend the proposed method in the field of open-set recognition for fault diagnosis.








\bibliographystyle{elsarticle-num} 
\bibliography{reference}

\end{document}